\documentclass{article}

 \usepackage[preprint]{neurips_2020} 

\usepackage[utf8]{inputenc} 
\usepackage[T1]{fontenc}    
\usepackage{hyperref}       
\usepackage{url}            
\usepackage{booktabs}       
\usepackage{amsfonts}       
\usepackage{nicefrac}       
\usepackage{microtype}      
\usepackage{dsfont}

\usepackage{graphicx}
\usepackage{amsmath}
\usepackage{amssymb}
\usepackage{adjustbox}
\usepackage{verbatim}
\graphicspath{ {figures/} }
\usepackage{multirow}
\usepackage[table,xcdraw]{xcolor}

\usepackage{tikz}
\usepackage{collcell}
\usepackage{etoolbox}

\newcommand*{\MinNumber}{0.0}%
\newcommand*{\MidNumber}{50.0} %
\newcommand*{\MaxNumber}{100.0}%

\newcommand{\gradientBLIMP}[1]{
    \ifdim #1 pt > \MidNumber pt
        \pgfmathparse{max(min(
        100.0*
        min(
        0.8*(#1 - \MidNumber)/((\MaxNumber-\MidNumber)/2), 
        0.5*(#1 - \MidNumber)/(\MaxNumber-\MidNumber)+0.5
        )
        ,100.0),0.00)} %
        \hspace{-0.33em}
        \xdef\tempa{\pgfmathresult}
        \cellcolor{black!35!green!\tempa!white!}{#1}
    \else
        \pgfmathparse{max(min(
        100.0*
        min(
        0.8*((\MaxNumber-\MidNumber) - #1)/((\MaxNumber-\MidNumber)/2),
        0.5*(\MidNumber - #1)/(\MidNumber-\MinNumber)+0.5
        )
        ,100.0),0.00)} %
        \hspace{-0.33em}
        \xdef\tempa{\pgfmathresult}
        \cellcolor{red!\tempa!white}{#1}
    \fi
 }

\newcommand*{\minval}{-5.0} %
\newcommand*{\midval}{0.} %
\newcommand*{\maxval}{9.0} %

\newcommand{\gradientBox}[1]{
    \ifdim #1 pt > \midval pt
        \pgfmathparse{max(min(
        100.0*(#1 - \midval)/(\maxval-\midval)
        ,100.0),0.00)} %
        \hspace{-0.33em}
        \xdef\tempa{\pgfmathresult}
        \cellcolor{black!35!green!\tempa!white!}{#1}
    \else
        \pgfmathparse{max(min(
        100.0*(\midval - #1)/(\midval-\minval)
        ,100.0),0.00)} %
        \hspace{-0.33em}
        \xdef\tempa{\pgfmathresult}
        \cellcolor{red!\tempa!white}{#1}
    \fi
 }
 
\newcommand*{\minvalABXw}{0.}%
\newcommand*{\midvalABXw}{10.95}%
\newcommand*{\maxvalABXw}{100.0}%
\newcommand{\gradientBoxABXw}[1]{
    \ifdim #1 pt > \midvalABXw pt
        \pgfmathparse{max(min(
        100.0*(#1 - \midvalABXw)/(\maxvalABXw-\midvalABXw)
        ,100.0),0.00)} %
        \hspace{-0.33em}
        \xdef\tempa{\pgfmathresult}
        \cellcolor{red!\tempa!white}{#1}
    \else
        \pgfmathparse{max(min(
        100.0*(\midvalABXw - #1)/(\midvalABXw-\minvalABXw)
        ,100.0),0.00)} %
        \hspace{-0.33em}
        \xdef\tempa{\pgfmathresult}
        \cellcolor{black!35!green!\tempa!white!}{#1}
    \fi
 }
 
\newcommand*{\minvalABXa}{0.}%
\newcommand*{\midvalABXa}{20.94}%
\newcommand*{\maxvalABXa}{100.0}%
\newcommand{\gradientBoxABXa}[1]{
    \ifdim #1 pt > \midvalABXa pt
        \pgfmathparse{max(min(
        100.0*(#1 - \midvalABXa)/(\maxvalABXa-\midvalABXa)
        ,100.0),0.00)} %
        \hspace{-0.33em}
        \xdef\tempa{\pgfmathresult}
        \cellcolor{red!\tempa!white}{#1}
    \else
        \pgfmathparse{max(min(
        100.0*(\midvalABXa - #1)/(\midvalABXa-\minvalABXa)
        ,100.0),0.00)} %
        \hspace{-0.33em}
        \xdef\tempa{\pgfmathresult}
        \cellcolor{black!35!green!\tempa!white!}{#1}
    \fi
 }
 
\newcommand{\mcrot}[4]{\multicolumn{#1}{#2}{\rlap{\rotatebox{#3}{#4}~}}} 

\newcommand\T{\rule{0pt}{3.1ex}}       
\newcommand\B{\rule[-1.7ex]{0pt}{0pt}} 

\makeatletter
\def\thickhline{%
  \noalign{\ifnum0=`}\fi\hrule \@height \thickarrayrulewidth \futurelet
   \reserved@a\@xthickhline}
\def\@xthickhline{\ifx\reserved@a\thickhline
               \vskip\doublerulesep
               \vskip-\thickarrayrulewidth
             \fi
      \ifnum0=`{\fi}}
\makeatother

\newlength{\thickarrayrulewidth}
\title{The Zero Resource Speech Benchmark 2021:\\ Metrics and baselines for unsupervised \\spoken language modeling}

\author{

  Tu Anh Nguyen\thanks{Equal contribution as first authors. $^{\dagger}$ Equal contributions as last authors.} \\
  Facebook AI Research \& EHESS, \\
  ENS-PSL Univ., CNRS, INRIA, France\\
  \texttt{\small nguyentuanh208@gmail.com} \\
    \And 
  Maureen de Seyssel$^{*}$ \\
  EHESS, ENS-PSL Univ., CNRS, INRIA\\
  \& U. Paris, France\\
  \texttt{\small maureen.deseyssel@gmail.com} \\
  \And 
  Patricia Rozé\\
  ENS-PSL Univ., CNRS\\
  France\\
  \texttt{\small patricia.roze@ens.fr} \\
  \And
  Morgane Rivière\\
  Facebook AI Research\\
  France\\
  \texttt{\small mriviere@fb.com} \\
  \And
  Evgeny Kharitonov\\
  Facebook AI Research\\
  France\\
  \texttt{\small kharitonov@fb.com} \\
  \And
  Alexei Baevski\\
  Facebook AI Research\\
  France\\
  \texttt{\small abaevski@fb.com} \\
  \And
  Ewan Dunbar$^{\dagger}$\\
  U. Paris Diderot, France\\
  \& U. Toronto, Canada\\
  \texttt{\small ewan.dunbar@utoronto.ca } \\
   \And
  Emmanuel Dupoux$^{\dagger}$\\
  Facebook AI Research \& EHESS, \\
  ENS-PSL, CNRS, INRIA, France\\
  \texttt{\small emmanuel.dupoux@gmail.com} \\
 
}

\begin{document}
\maketitle
\begin{abstract}

We introduce a new unsupervised task, spoken language modeling: the learning of linguistic representations from raw audio signals without any labels, along with %
the Zero Resource Speech Benchmark 2021: a suite of 4 black-box, zero-shot metrics probing for the quality of the learned models at 4 linguistic levels: phonetics, lexicon, syntax and semantics. We present the results and analyses of a composite baseline made of the concatenation of three unsupervised systems: self-supervised contrastive representation learning (CPC), clustering (k-means) and language modeling (LSTM or BERT). The language models learn on the basis of the pseudo-text derived from clustering the learned representations. This simple pipeline shows better than chance performance on all four metrics, demonstrating the feasibility of spoken language modeling from raw speech. It also yields worse performance compared to text-based `topline' systems trained on the same data, delineating the space to be explored by more sophisticated end-to-end models.  


\end{abstract}

%


\vspace{-1em}
\section{Introduction}
\vspace{-0.7em}

In recent work, self-supervised techniques from vision and NLP have been applied to large datasets of raw audio, giving rise to very effective methods of pretraining for downstream ASR tasks, particularly in the low resource scenario \citep{schneider2019wav2vec,baevski2019effectiveness,chung2019generative,baevski2020wav2vec,riviere2020unsupervised,kawakami2020,wang2020}.
The approaches based on transformers and masking objectives, strikingly similar to the models used to train language models, are especially intriguing. The fact that these approaches yield excellent ASR performance (less than 10\% WER) with as little as 10 minutes of labels plus a language model (LM), or with 10 hours of labels but no LM \citep{baevski2020wav2vec}, suggests that these systems may actually go beyond acoustic modeling, learning their own LM from raw audio. Such work therefore connects with research into the \textit{zero resource} setting, which aims at  learning linguistic representations from scratch for language with little or no textual resources. However, up to now, there exists no established benchmark to analyse the representations learned by such models beyond the acoustic/phonetic level. 


Typically, language models trained from text are evaluated using scores like perplexity. Unfortunately, this simple approach cannot be used here, since perplexity scores computed from learned discrete units  
vary according to granularity, making model comparison impossible. This is why we chose to follow a black-box NLP strategy: our metrics do require expert linguistic labels for the dev and test sets, but are \textit{zero-shot} in that they do not require training a classifier, they use \textit{simple tasks} enabling direct human/machine comparison, and they give \textit{interpretable} scores at each linguistic level.
As seen in Table \ref{tab:metrics}, they can be divided into two types: distance-based and probability-based metrics. Distance-based metrics require models to provide a pseudo-distance computed over pairs of embeddings. The ABX score \citep{Schatz2013ABX}, already used for the evaluation of \textit{acoustic/phonetic} representations, falls in this category and provides a measure of how well separated phonetic categories are in a given embedding space. Here, we use the ABX score developed in Libri-light \citep{Kahn_2020_librilight}. Distance-based methods can also be used to evaluate the \textit{semantic} representation of words, by computing the correlation between these distances and human semantic similarity judgements \citep[see][]{schnabel2015evaluation,faruqui2016problems}. \citet{chung2018speech2vec} adapted this metric to speech, which we compiled into our sSIMI dataset. Probability-based metrics require models to compute a pseudo-probability for a given test input (non-normalized non-negative number for a given input waveform). The pseudo-probabilities are computed over pairs of inputs, one of which is acceptable in the tested language and the other not. Such methods have been used in NLP to evaluate the \textit{syntactic} abilities of language models, by comparing the probabilities of grammatical versus  ungrammatical sentences\citep{warstadt2019blimp}, and we built the sBLIMP dataset upon this work. Finally, in our sWUGGY dataset, we extend this logic to the \textit{lexical} level by comparing the pseudo-probability associated to words and nonwords. The four metrics are presented in more details in Section \ref{sec:metrics}.

Next, we apply these metrics to a simple baseline system (Section \ref{sec:models}), built on contrastive pretraining \cite[Contrastive Predictive Coding, CPC,][]{oord2018cpc,riviere2020unsupervised}, followed by k-means clustering, which we use to decode a speech dataset \citep[LibriSpeech,][]{Panayotov2015librispeech} into pseudo-text. This pseudo-text is used to train a language model varying in compute budget: an LSTM (smaller budget) or BERT (larger budget) model. We show (Section \ref{sec:results}) that such simple baseline models give better than chance performance on all 4 metrics, demonstrating that it has learned representations at the four corresponding linguistic levels. However, comparison with a text-based BERT topline system trained on the phonetic transcription of the same training data shows that the speech input raises challenges for the LM component of the model that need to be addressed in further work. Datasets and baselines will be open sourced to encourage bridging the gap between speech and text-based systems.  


\begin{table*}[]
\def\wexample{4.6cm}
\label{tab:metrics}
\caption{\textbf{Summary description of the four Zero Resouce Benchmark 2021 metrics}. The metrics in light blue use a pseudo-distance $d$ between embeddings ($d_h$ being from human judgments), the metrics in light orange use a pseudo-probability $p$ computed over the entire input sequence.}
\begin{tabular}{p{1.3cm}p{1.6cm}lp{2.8cm}p{\wexample}}
\hline
Linguistic \mbox{level}   & Metrics & Dataset& Task  &Example \\
\hline
\rowcolor[HTML]{E7FAFE} 
acoustic-phonetic  & ABX              &Libri-light& $d(a,x) < d(b,x)?$ \mbox{$a \in A, b\in B,$} \mbox{$x\neq a \in A$} & \parbox[t]{\wexample}{
within-speaker: $(\textrm{apa}_{s_1},\textrm{aba}_{s_1},\textrm{apa}_{s_1})$\\ across-speaker: $(\textrm{apa}_{s_1},\textrm{aba}_{s_1},\textrm{apa}_{s_2})$} \\
\rowcolor[HTML]{FFF7E5} 
lexicon            & spot-the-word    &sWUGGY   & $p(a)\textgreater{}p(b)?$  & \parbox[t]{\wexample}{(brick, blick)\\(squalled, squilled)}\\
\rowcolor[HTML]{E7FAFE} 
lexical \mbox{semantics}  & similarity judgement &sSIMI & $d(a,b) \propto d_{h}(a,b) ?$ &\parbox[t]{\wexample}{ (abduct, kidnap) : 8.63\\ (abduct, tap) : 0.5 }\\
\rowcolor[HTML]{FFF7E5} 
\rowcolor[HTML]{FFF7E5} 
syntax             & acceptability judgment &sBLIMP& $p(a) > p(b) ?$ & \parbox[t]{\wexample}{(dogs eat meat, dogs eats meat)\\ (the boy can't help himself, the boy can't help herself)} \\
\rowcolor[HTML]{E7FAFE} 
\hline
\end{tabular}
\vspace{-1em}
\end{table*}

\vspace{-1em}
\section{Related work}
\vspace{-0.9em}

%

\paragraph{Zero Resource Speech Challenge Series.} Previous work \citep{versteegh2015zero, dunbar2017zero, dunbar2019zero,dunbar2020} has focused on establishing benchmarks for unsupervised learning of an entire dialogue system, but has so far remained at a rather low level (acoustic, lexical). Acoustic modeling has used two metrics: ABX, a distance-based metric to be discussed later, and opinion scores on TTS output (whereby the discovered units are used to resynthesize speech). As for the lexical level, past work has focused on using the NLP metrics developed for word segmentation \citep{ludusan2014bridging}. However, these metrics assume that the models should discover words explicitly. The success of character-based language models suggests that it is possible to learn high-level linguistic concepts without explicitly segmenting words \citep[see][]{hahn2019}. 

\paragraph{Black box NLP.}
Among the variety of black-box linguistic tasks, psycholinguistically-inspired ones enable direct comparison of models and humans.  Grammaticality judgments for recurrent networks have been investigated since \citet{allen1999:emergence}, who use closely matched pairs of sentences to investigate grammatical correctness. This approach has recently been adopted to assess the abilities of RNNs, and LSTMs in particular, in capturing syntactic structures. For instance, \citet{linzen:2016} and \citet{gulordava:2018} use word probes in minimally different pairs of English sentences to study number agreement. To discriminate grammatical sentences from ungrammatical ones, they retrieve the probabilities of the possible morphological forms of a target word, given the probability of the previous words in the sentence. Practically, in the sentence ``the boy \underline{is} sleeping'', they assume the network has detected number agreement if \textit{$\mathbf{P}(w = is) > \mathbf{P}(w  = are)$}. This methodology has also been adapted by \citet{goldberg:2019} to models trained with a masked language-modeling objective. 
Similarly, \citet{ravfogel2018can} use word probes to examine whether LSTMs understand Basque agreement and \cite{godais2017charnlm} to test the lexical level in character-based LM. 

\vspace{-1em}
\section{Methods}\label{sec:method}
\vspace{-.5em}

\subsection{Training set}\label{sec:data} 
\vspace{-.5em}

We used as a training set the LibriSpeech 960h dataset \citep{Panayotov2015librispeech}. We also included in this work the clean-6k version of the Libri-light dataset \citep{Kahn_2020_librilight} which is a huge collection of speech for unsupervised learning. A phonetic transcription of the LibriSpeech dataset was also employed. To obtain this, we used the original LibriSpeech lexicon, as well as the G2P-seq2seq toolkit\footnote{https://github.com/cmusphinx/g2p-seq2seq} to generate the phonetic transcriptions of words lacking from the lexicon. We generated a forced-alignment version of Librispeech using the abkhazia library\footnote{https://github.com/bootphon/abkhazia}. This enabled us to provide comparative text-based topline systems along with the speech baseline. 


\vspace{-.5em}

\subsection{Metrics}\label{sec:metrics}
\vspace{-.5em}

We set up four metrics with their accompanying datasets, to evaluate the sLMs at four levels: phonetic (the Libri-light ABX metrics), lexical (the sWUGGY spot-the-word metrics), syntactic (the sBLIMP acceptability metrics) and semantic (the sSIMI similarity metric). The 4 datasets are composed of speech sounds extracted from LibriSpeech (sSIMI), or synthetic stimuli constructed with the Google API\footnote{https://cloud.google.com/text-to-speech} using 4 different voices, two males and two females (sWUGGY, sBLIMP, sSIMI)\footnote{We use WaveNet voices A, C, D and F. All dev set stimuli are synthesised in all four voices. Stimuli in the sSIMI and sBLIMP test sets are split evenly among the four different voices, and sWUGGY uses all four for each test set stimulus.}. When synthetic, the stimuli were subsequently force-aligned to retrieve the phonetic boundaries. The datasets containing words or sentences were filtered to only contain the LibriSpeech vocabulary, and are split into dev and test sets. 

\paragraph{Phonetics: Libri-light ABX metrics.}

The ABX metric consists in computing, for a given contrast between two speech categories $A$ and $B$ (e.g., the contrast between triphones `aba' and `apa'), the probability that two sounds belonging to the same category are closer to one another than two sounds that belong to different categories. Formally, we compute an asymmetric score, with $a$ and $x$, different tokens belonging to category $A$ (of cardinality $n_A$) and $b$ belonging to $B$ ($n_B$), respectively:  

\vspace{-0.9em}

\begin{equation}
\hat{e}(A, B) := \frac{1}{ n_{A}(n_A - 1) n_B}\sum_{\substack{a,x \in A \\ x \neq a}}\sum_{b \in B } \left[\mathds{1}_{d(b, x) < d(a, x)}\right.
+ \left. \frac{1}{2}\mathds{1}_{d(b, x) = d(a, x)}\right]
\end{equation}

The score is symmetrized and aggregated across all minimal pairs of triphones like `aba', `apa', where the change only occurs in the middle phoneme. This score can be computed within speaker (in which case, all stimuli $a$, $b$ and $x$ are uttered by the same speaker) or across speaker ($a$ and $b$ are from the same speaker, and $x$ from a different speaker). This score requires a pseudo-distance between acoustic tokens computed by  averaging along a dynamic time warping path a framewise distance (KL or angular distance). This metric is agnostic to the dimensionality of the embeddings, can work with discrete or continuous codes, and has been used to compare ASR speech features \citep{schatz2016abx}.  Here, we run this metric on the pre-existing Libri-light dev and test sets, which has been already used to evaluate several self-supervised models \citep{Kahn_2020_librilight,riviere2020unsupervised}.

\paragraph{Lexicon: sWUGGY spot-the-word metrics.} 
We built on \citet{godais2017charnlm} which used the `spot-the-word' task. In this task, networks are presented with a pair of items, an existing word and a matching nonword, and are evaluated on their capacity to attribute a higher probability to the existing word. The spot-the-word metric corresponds to the average accuracy of classifying the words and nonwords correctly across each pair.

The nonwords are produced with WUGGY\citep{keuleers2010wuggy}, which generates for a given word, a list of candidate nonwords best matched in phonotactics and syllabic structure.
Because we were aiming at speech stimuli, we needed additional constraints to ensure that (i) the audio synthesis of the pairs would be of good quality, and (ii) that the pairs would have matching unigram and bigram scores relative to their phonemes. On a sample of 100 word/nonword pairs, and with feedback from a native English speaker informant, we designed a synthesis-quality rule. The rule consists of testing whether the original phonetic transcription matches the output of a back-to-back phoneme-to-grapheme (p2g) and grapheme-to-phoneme encoding (g2p).\footnote{We used the G2P-seq2seq toolkit.} Only pairs where both the words and nonwords passed this test were kept. 
We added additional constraints using a stochastic sampler to also match unigram and bigram phoneme frequencies (see Supplementary Material \ref{SM:sampling}). 
The final sWUGGY test and development sets consists of 20,000 and 5,000 pairs respectively, with the existing words being part of the LibriSpeech train vocabulary. We also prepared additional OOV-sWUGGY test and development sets consisting of 20,000 and 5,000 pairs respectively, with existing words which do not appear in the LibriSpeech training set.


The spot-the-word accuracy is the average of the indicator function $1_{PP(word_k)>PP(nonword_k)}$ over the set of pairs ${(word_k,nonword_k)}$, where \textit{PP} is a pseudo-probability (a possibly non-normalized non-negative number) assigned to each input file by the model.

\paragraph{Syntax: sBLIMP acceptability metrics.}  

This part of the benchmark is adapted from BLIMP \citep{warstadt2019blimp}, a dataset of linguistic minimal sentence pairs of matched grammatical and ungrammatical sentences. Similarly to the preceding test, the task is to decide which of the two members of the pair is grammatical based on the probability of the sentence. 
We adapted the code used to generate the BLIMP dataset \citep{warstadt2019blimp} in order to create sBLIMP, specifically tailored for speech purposes. In BLIMP, sentences are divided into twelve broad categories of syntactic paradigms. These categories are themselves divided into 68 specific paradigms containing 1000 sentence pairs each, automatically generated using an expert hand-crafted grammar (this includes an additional subcategory which was added to the code subsequent to \citet{warstadt2019blimp}.

To make this dataset `speech-ready,' we discarded five subcategories and slightly modified the grammar for nine additional subcategories in order to ensure sentences had appropriate prosodic contours. We also removed from the vocabulary all words absent from the LibriSpeech train set \citep{Panayotov2015librispeech}, as well as compound words and homophones that could cause further comprehension issues once synthesised. 5000 sentence pairs were then generated for each of the 63 remaining subcategories. 
We sampled sentence pairs from the generated pool to create a development and a test set, ensuring that the larger linguistic categories were sampled so as to balance the n-gram language model scores (see Supplementary Material \ref{SM:sampling}). The test and development sets contain 63,000 and 6,300 sentence pairs respectively, with no overlap in sentence pairs. Stimuli were then synthesized and force-aligned as described at the beginning of the section.

Similar to the spot-the-word metric, the acceptability judgment metric requires a pseudo-probability for each given input file. The sentence acceptability accuracy is reported similarly to the spot-the-word accuracy with the pairs of
grammatical and ungrammatical sentences in the sBLIMP dataset.



\paragraph{Lexical semantics: sSIMI similarity metrics.} Here, the task is to compute the similarity of the representations of pairs of words and compare it to human similarity judgements. 
Based on previous work \citep{chung2018speech2vec}, we used a set of 13 existing semantic similarity and relatedness tests to construct our similarity benchmark.
The similarity-based datasets include WordSim-353 \citep{yang2006verb},
WordSim-353-SIM \citep{agirre2009study}, mc-30 \citep{miller1991contextual}, rg-65 \cite{rubenstein1965contextual}, Rare-Word (or rw) \citep{luong2013better}, simLex999 \citep{hill2015simlex},
simverb-3500 \citep{gerz2016simverb}, verb-143 \citep{baker2014unsupervised} , YP-130 \cite{yang2006verb}
and the relatedness-based datasets include MEN \citep{bruni2012distributional}, Wordsim-353-REL \citep{agirre2009study}, mturk-287 \citep{radinsky2011word}, mturk-771 \citep{halawi2012large}.
All scores were normalised on a 0-10 scale, and pairs within the same dataset containing the same pair of words but in the opposite order were averaged.   Pairs containing a word not in the LibriSpeech train set \cite{Panayotov2015librispeech} were discarded.

We  selected as a development set the mturk-771 dataset, which was, in  preliminary study using character- and word-based LMs, both highly correlated with all other datasets and was large enough to be used as a development set. It was also ensured that no pair from the development set was present in any of the test sets. All other twelve datasets were used as test sets. 
We then created two subsets of audio files, one synthetic, one natural.
For the first, we followed the synthesis and forced alignment procedures described at the beginning of the section. For the second, we retrieved the audio extracts from LibriSpeech corresponding to each word, following the process presented in \citep{chung2018speech2vec}. The natural subset is therefore smaller than its synthesized counterpart as we had to discard pairs from the test and dev sets which were not present in the LibriSpeech test and dev sets respectively. However, in this natural subset, each word may appear in multiple tokens, providing phonetic diversity; duplicated scores are averaged in the analysis step. The synthesised subset is composed of 9744 and 705 word pairs for the test and dev sets respectively, and the LibriSpeech subset is composed of 3753 and 309 pairs for the test and dev sets.

The semantic similarity score is reported as the Spearman's rank correlation coefficient $\rho$ between the semantic distance scores
given by the model and the true human scores in the dataset. Note that in this work all the semantic similarity scores are multiplied by 100 for clarity.


\subsection{Models}\label{sec:models}
\paragraph{Baseline models.}
Our baseline models are a composite of three components: an acoustic model (CPC), a clustering module (k-means) and a language model (LSTM or BERT) varying in size.

The acoustic model is built upon Contrastive Predictive Coding (CPC, \cite{oord2018cpc}), where the representation of the audio is learned by predicting the future through an autoregressive model. 
In more detail, given an input signal $\textbf{x}$, the CPC model embeds $\textbf{x}$ to a sequence of embeddings $\textbf{z}=(z_1,\dots,z_T)$ at a given rate through a non-linear encoder $g_{\text{enc}}$.
At each time step $t$, the autoregressive model $g_{\text{ar}}$ takes as input the available embeddings $z_1,\dots,z_t$ and produces a context latent representation $c_t=g_{\text{ar}}\left(z_1,\dots,z_t\right)$.
Given the context $c_t$, the CPC model tries to predict the $K$ next future embeddings $\{z_{t+k}\}_{1\leq k\leq K}$ by minimizing the following constrastive loss:
\begin{equation}\label{eq:CPCloss}
    \mathcal{L}_t = - \frac{1}{K}\sum_{k=1}^K\log \left[\frac{\exp \left(z_{t+k}^\top W_kc_t\right)}{\sum_{\Tilde{z}\in\mathcal{N}_t}\exp \left(\Tilde{z}^\top W_kc_t\right)}\right]
\end{equation}
where $\mathcal{N}_t$ is a random subset of negative embedding samples, and $W_k$ is a linear classifier used to predict the future $k$-step observation. 
We used a PyTorch implementation of CPC\footnote{https://github.com/facebookresearch/CPC\_audio} \citep{riviere2020unsupervised}, which is a modified version of the CPC model that stabilizes the CPC training by replacing batch normalization with a channel-wise normalization and improves the CPC model by replacing the linear classifier $W_k$ in equation (\ref{eq:CPCloss}) with a 1-layer Transformer network \citep{vaswani2017attention}. The encoder $g_{\text{enc}}$ is a 5-layer 1D-convolutional network with kernel sizes of 10,8,4,4,4 and stride sizes of 5,4,2,2,2 respectively, resulting in a downsampling factor of 160, meaning that the embeddings have a rate of 100Hz. The autoregressive model $g_{\text{ar}}$ is a multi-layer LSTM network, with the same hidden dimension as the encoder. 
For this baseline, we trained two different versions of CPC: CPC-small and CPC-big. Details are given in Table \ref{tab:CPCmodels}.

\begin{table}[t]
    \centering
    \caption{\textbf{Characteristics of the baseline acoustic CPC models}. We took the last LSTM layer of CPC-small and the second LSTM hidden layer of CPC-big as inputs to the clustering as they give the best ABX scores (Supplementary Table \ref{tab:gruABX}).}
    \label{tab:CPCmodels}
    \begin{tabular}{lcccc}
        \thickhline
        \rowcolor{white}
        & \multicolumn{2}{c}{CPC configuration} & \multirow{2}{*}{Training data} & \multirow{2}{*}{Input to kmeans}\\
        \cline{2-3}
        Model & autoregressive & hidden units & & \\
        \thickhline
        \rule{0pt}{2ex}CPC-small & 2-layer LSTM & 256 & LibriSpeech clean-100 & LSTM level 2 \\
        CPC-big & 4-layer LSTM & 512 & Libri-light clean-6k & LSTM level 2 \\
        \thickhline
    \end{tabular}
\end{table}

\begin{table}[t]
    \centering
    \caption{\textbf{Characteristics of the baseline LMs.} L refers to the number of hidden layers; ED, HD and FFD refer to the dimension of the embedding layer, hidden layer, and feed-forward output layer respectively; H refers to the number of attention heads in the BERT case.}
    \label{tab:LMarchi}
    \begin{tabular}{lcccccccc}
        \thickhline
        \rowcolor{white}
        & \multicolumn{5}{c}{Architecture} & nb & Train & Compute\\
        \cline{2-6}
        Model & L & ED & HD & FFD & H & parameters & data & Budget\\
        \thickhline
        \rule{0pt}{2ex}BERT & 12 & 768 & 768 & 3072 & 12 & 90M & LS960 & 48h - 32 GPUs \\
        BERT-small & 8 & 512 & 512 & 2048 & 8 & 28M& LS960& 60h - 1GPU \\
        LSTM & 3 & 200 & 1024 & 200 & - & 22M & LS960 & 60h- 1GPU\\
        \thickhline
    \end{tabular}
\end{table}

After training the CPC model, we then train a k-means clustering module on the outputs of either the final layer or a hidden layer of the autoregressive model. The clustering is done on the collection of all the output features at every time step of all the audio files in a given training set. After training the k-means clustering, each feature is then assigned to a cluster, and each audio file can then be discretized to a sequence of discrete units corresponding to the  assigned clusters.
The k-means training was done on the subset of LibriSpeech containing 100 hours of clean speech.



Finally, with the discretized version of the audio files, we train language models on the discretized units. We establish two `low budget' and two `high budget' baselines, based on the number of parameters and the compute resources necessary to train them. The high budget used a BERT-based architecture \citep{devlin2018bert} trained either on CPC-small or CPC-big plus k-means-50 pretrained units. The low budget architectures were a two-layer LSTM and a small BERT architecture (see Table \ref{tab:LMarchi} for details); they both used the units from the CPC-big pretraining. 
Following \cite{Baevski2020vq-wav2vec}, we trained the BERT models with only the masked token prediction objective. We also followed \cite{Baevski2020vq-wav2vec} by masking a span of tokens in the input sequence instead of a single token (otherwise the prediction would be trivial to the model as discretized units tend to replicate). We masked $M$ consecutive tokens for each span, where $M\sim \mathcal{N}(10,10)$, with a total masking coverage of roughly half of the input tokens (spans may overlap). All models were trained on LibriSpeech 960h.
The BERT models were trained with a total batch size of 524k tokens, and the LSTM model was trained with a total batch size of 163k tokens. The learning rate was warmed up to a peak value of $1\times 10^{-5}$. All the implementation was done via fairseq \citep{ott-etal-2019-fairseq}.

\vspace{-0.7em}

\paragraph{The Topline models. } 
For topline comparison, we trained a BERT model on force-aligned phonemes using the gold transcription of the LibriSpeech dataset. We also employed the span masking similarly to the baseline model. 
In addition to the BERT trained on forced alignments, we also included a BERT model trained on the gold phonetic transcription of the LibriSpeech dataset, with the difference that we only mask one token instead of a span of tokens. 
For an absolute topline comparison, we used the pretrained RoBERTa large model \citep{Liu2019RoBERTa}, which was trained on 50K subword units on a huge dataset of total 160GB, 3000 times bigger than the transcription of the LibriSpeech 960h dataset.

\vspace{-1em}
\section{Results}\label{sec:results}
\vspace{-0.7em}
\subsection{Libri-light ABX}
\vspace{-0.7em}

\paragraph{Computing distances. }
We used the average angular distance (arccos of the normalized dot product) of the representations along the DTW-realigned path, as used by default in previous challenges \citep{versteegh2015zero, dunbar2017zero, dunbar2019zero}.  For our baseline models, we computed the ABX scores over one-hot representations of discretized units of the audio files.
\vspace{-0.7em}

\paragraph{Results.}
\begin{table}[]
\caption{\textbf{Within and Across Speaker ABX error} (lower is better) on Libri-light dev-clean and -other for two unsupervised models, before and after clustering (1-hot representations).}
\label{tab:ABX}
\begin{center}
\begin{tabular}{lcclcc}
\thickhline
\rowcolor{white}
 & \multicolumn{2}{c}{within} & & \multicolumn{2}{c}{across}  \\
 \cline{2-3} \cline{5-6}
Embedding\rule{0pt}{1ex}        & dev-clean  & dev-other   & & dev-clean & dev-other \\
\thickhline
\rule{0pt}{2ex}MFCC & 10.95  & 13.55 &&  20.94 & 29.4\\
\hline
CPC-small    & 6.24     & 8.48     & & 8.17     & 13.55  \\
\ \ +kmeans-50   & 10.26    & 14.24    & & 14.17    & 21.26  \\
CPC-big      & 3.41     & 4.85     & & 4.18     & 7.64  \\
\ \ +kmeans-50\rule[-1.2ex]{0pt}{0pt}   & 6.38     & 10.22    & & 8.26    & 14.86  \\
\thickhline
\end{tabular}
\end{center}
\vspace{-2em}

\end{table}

We first ran experiments varying the number of clusters. As seen in Supplementary Table \ref{tab:kmeansABX}, too few or too many clusters gives rise to worse ABX performance, with a sweet spot at 50 clusters, which is the number we retain for the remainder of the paper. In Table \ref{tab:ABX}, we present the result of the ABX for our two models (CPC-small and CPC-big), before and after clustering. One can see that the CPC-big model yields better performance than the CPC-small model (we retain the big model for the rest of the experiments), and the clustering step yields an increase in error of between 60-100\%. Still, the performances are better than for an MFCC representation, with a much more compact code.

\vspace{-0.7em}

\subsection{sWUGGY spot-the-word}\label{sec:WUGGY}
\vspace{-0.7em}
\paragraph{Computing the pseudo-probability.}
Given an audio file $x$, we first discretize $x$ into a sequence of discretized units $q_1...q_T$. 
Then, following \cite{salazar-etal-2020-masked}, we propose the following pseudo-probability score for our BERT models trained with a span-masked token prediction objective:
\begin{equation*}\label{eq:span-PP}
    \text{span-PP}_{M_d, \Delta t}(q_{1}..q_{T})=\prod_{\substack{i=1+j\Delta t\\\lfloor(T-1)/\Delta t\rfloor \geq j\geq 0}} P(q_{i}..q_{i+M_d}|q_{1}..q_{i-1}q_{i+M_d+1}..q_{T}),
\end{equation*}
where $M_d$ is a chosen decoding span size, and $\Delta t$ is a temporal sliding size. 
For the LSTM model, we computed the probability of the discretized sequence with the classic left-to-right scoring style obtained by the chain rule: $P(q_{1}..q_{T})=\prod_{i=1}^T P(q_{i}|q_{1}..q_{i-1})$.

\paragraph{Results.}
We determined the optimal masking (Supplementary Table \ref{tab:masking}) to be $\Delta t = 5$ and $M_d =15$. We kept this setting for all other experiments involving pseudo-probabilities. Table \ref{tab:overall} presents the average of the four baseline systems and in Figure \ref{fig:WUGGY}, the detailed performances of the baseline compared to n-gram controls and toplines. The performance of all four baselines is consistently better than chance and n-gram controls.

\vspace{-0.7em}
\subsection{sBLIMP acceptability}
\vspace{-0.7em}
\paragraph{Computing the pseudo-probability.} We computed the pseudo-probability as in Section \ref{sec:WUGGY}. 

\vspace{-0.7em}
\paragraph{Results.}
The aggregate results are shown in Table \ref{tab:overall} and the detailed ones on the best system in Table \ref{tab:BLIMP}. The results of this test, while above chance are considerably lower than the text-based toplines.  

\subsection{sSIMI semantic similarity}
\paragraph{Computing the distance.} We computed the semantic distance between two audio files $x$ and $y$ as the similarity between the two corresponding discretized sequences $q^x_1...q^x_T$ and $q^y_1...q^y_S$. To obtain this, we extracted outputs from a hidden layer of the LM to the two discretized sequences, aggregating them with a pooling function to produce a fixed-length representation vector for each sequence, and computed the cosine similarity between the two representation vectors:
\begin{equation*}
    d_{SEM}(x, y) = sim\left(f_{pool}\left(h^{(i)}(q^x_1...q^x_T)\right), f_{pool}\left(h^{(i)}(q^y_1...q^y_S)\right)\right),
\end{equation*}
where $f_{pool}$ is the pooling function and $h^{(i)}(\cdot)$ is the output of the $i^{th}$ hidden layer of the LM.

As each word consists of possibly several voices, we averaged the similarity distance over pairs of the same voice for the synthetic subset, and all possible pairs for the LibriSpeech subset.


\vspace{-0.7em}
\paragraph{Results.} For each model, we chose the pooling function and the hidden level that give the best score on the dev set, and computed the score on the corresponding test set. The aggregate results are in Table \ref{tab:overall}, and a detailed layer-by-layer analysis in Table \ref{tab:ABX-and-SIMI}. The scores for semantic similarity are overall modest, compared to BERT systems trained on larger units (BPE). However, one can observe that the best layers for semantic similarity occur towards the first third of the transformer, and that max pooling seems to be best. This contrasts with the best layers for acoustic similarity (as indexed by ABX), which occur at the extremities.

\vspace{-0.7em}
\subsection{Model comparison}
\vspace{-0.7em}

 The overall results are in Table \ref{tab:overall}. They show that the four baseline models are above chance in the four tasks, even low budget ones, although there is substantial variation between tasks. While task at the lexical level is substantially above chance, the syntactic and semantic tasks show room for improvement compared to text-based toplines trained on similar amounts of data. 
\vspace{-0.7em}
\section{Discussion}
\vspace{-0.7em}
\begin{table}[ht]
\caption{Overall performance of our baseline and topline models on dev and test sets on our four zero-shot metrics. 
For baseline models, 
the k-means training (k=50) was performed on LibriSpeech clean-100h, and the LSTM/BERT models was trained on discretized units of LibriSpeech 960h. For topline comparisons, we included a BERT model trained on the forced aligned frames of LibriSpeech 960h, a BERT model trained on the gold phonetic transcription of LibriSpeech 960h, and a RoBERTa large model pretrained on a text dataset 3000 times bigger than the transcription of LibriSpeech 960h.}
\label{tab:overall}
\vspace{-1em}
\begin{center}
\begin{adjustbox}{max width=0.97\textwidth, center}
\begin{tabular}{l @{\hspace{0.5\tabcolsep}}c@{\hspace{0.7\tabcolsep}} c@{\hspace{0.5\tabcolsep}}c@{\hspace{0.5\tabcolsep}}c@{\hspace{0.5\tabcolsep}}c @{\hspace{0.9\tabcolsep}} c@{\hspace{0.9\tabcolsep}}c @{\hspace{0.9\tabcolsep}}@{\hspace{0.9\tabcolsep}} c @{\hspace{0.5\tabcolsep}} c @{\hspace{0.9\tabcolsep}} }
\thickhline
\rowcolor{white}
\rule{0pt}{2ex}  &  & \multicolumn{2}{c}{ABX within} & \multicolumn{2}{c}{ABX across} & sWUGGY & sBLIMP & \multicolumn{2}{c}{sSIMI} \\
 \hline
System\rule{0pt}{1ex}    & Set     & clean & other & clean & other &  &  & synth. & libri.   \\
\thickhline
\multicolumn{10}{c}{\textit{Low budget baseline systems}}  \\
\multirow{2}{*}{CPC-big+km50+BERT-small} & dev & 6.38 &	10.22  &	8.26 &	14.86 &	65.81 &	52.91 & 3.88 &	5.56 \\
& test & 6.71 &	10.62 &	8.41 &	15.06 &	65.94 &	 53.02 & 3.02 & 0.06 \\
\multirow{2}{*}{CPC-big+km50+LSTM} & dev & 6.38 &	10.22 &	8.26 &	14.86 &	66.13  &	53.32 &	4.42 &	7.56 \\
& test  &	6.71 &	10.62 &	8.41 &	15.06 &	66.22 &		52.89  & 7.35 &	6.66\\
\hline
\multicolumn{10}{c}{\textit{High budget baseline systems}}\\
\multirow{2}{*}{CPC-small+km50+BERT} &dev &  10.26 &	14.24 &	14.17 &	21.26  &	70.69  &	54.26  & 2.99 &	6.68 \\
& test &  	10.07 &	14.71 &	13.45 &	22.42  &	70.50  & 54.61  &	8.96 & -1.55\\
\multirow{2}{*}{CPC-big+km50+BERT} & dev & 6.38 &	10.22  &	8.26 &	14.86 &		75.56  &	56.14  & 6.25 &	8.72 \\
& test & 	6.71 &	10.62 &	8.41 &	15.06  &	75.51  &	56.16  & 5.17 &	1.75\\
\hline
\multicolumn{9}{c}{\textit{Topline systems}}\\
\multirow{2}{*}{Forced align BERT} & dev & 0.00&	0.00&	0.00&	0.00&	92.19&	63.72&	7.92 & 4.54\\
& test &	0.00&	0.00&	0.00&	0.00&		91.88&	63.16&		8.52 & 2.41\\
\multirow{2}{*}{Phone BERT} &dev & 	-&	-&	-&	-&	97.90&		66.78&	9.86 & 16.11 \\
&test &  - &	-&	-&	-&		97.67&	66.91&	12.23 & 20.16\\
\multirow{2}{*}{RoBERTa large} & dev &	- &	- &	- &	- &	96.58 &	81.56&		32.28 & 28.96\\
& test & - &	- &	- &	-  &	96.25&		82.11&	33.16 & 27.82\\
\thickhline
\end{tabular}
\end{adjustbox}
\end{center}
\vspace{-1em}
\end{table}

We introduced the new Zero Resource Speech Benchmark 2021 for spoken language models. It is composed of 4 zero-shot tests probing 4 linguistic levels: acoustic, lexical, syntactic and semantic. We showed that a simple CPC+clustering+LM trained on LibriSpeech can perform above chance on all of these tests, outperforming n-gram models, while being worse than text-based models trained on the same data. This shows both that the spoken LM task is feasible, and that there is room for improvement. 

Obvious directions for research include improving the representation learning component, the clustering methods, and the transformer, which have not been particularly tuned for this benchmark. There are also end-to-end models like wav2vec \citep{baevski2020wav2vec} and other masking systems \citep{wang2020} that could be tried in this context. The performance gap between the RoBERTa large system and our toplines trained on LibriSpeech suggest that much is to be gained by increasing the size of the training set, which can be obtained by large unlabelled audio datasets like LibriVox. Finally, even though this benchmark is intended for developing speech technology for low resource languages, significant resources are still required to construct the test sets and metrics (phonetic dictionary, aligned speech, grammar, TTS or trained speakers to make the stimuli). More work is needed to reduce this footprint and scale up this benchmark to languages other than English.

\vspace{-0.7em}
\section*{Broader Impact}
\vspace{-0.7em}

The metrics developed here may help improve interpretability of unsupervised systems. Research within the Zero Resource setting may help for developing speech technology for low resourced languages, or for languages with no textual resources, which cannot be addressed in the supervised setting. Even for high resource languages, learning a language model from raw speech would help address dialect variation, including minorities, making speech technology more inclusive. Broadening the reach of speech technology might be used to increase the economic dominance of already-large actors if developed with proprietary resources. To minimize this, we engage the community through an open source benchmark.


\section*{Acknowledgments}
The work for MS, PR and for EDupoux and TAN in their EHESS role was supported by the Agence Nationale de la Recherche (ANR-17-EURE-0017 Frontcog, ANR-10-IDEX-0001-02 PSL*, ANR-19-P3IA-0001 PRAIRIE 3IA Institute) and grants from CIFAR (Learning in Minds and Brains) and Facebook AI Research (Research Grant). The work for EDunbar was supported by a Google Faculty Research Award and by the Agence Nationale de la Recherche  (ANR-17-CE28-0009 GEOMPHON,  ANR-18-IDEX-0001 U de Paris, ANR-10-LABX-0083 EFL).
\bibliography{main_bib}

\begin{thebibliography}{46}
\expandafter\ifx\csname natexlab\endcsname\relax\def\natexlab#1{#1}\fi

\bibitem[{Agirre et~al.(2009)Agirre, Alfonseca, Hall, Kravalova, Pasca, and
  Soroa}]{agirre2009study}
Eneko Agirre, Enrique Alfonseca, Keith Hall, Jana Kravalova, Marius Pasca, and
  Aitor Soroa. 2009.
\newblock A study on similarity and relatedness using distributional and
  wordnet-based approaches.

\bibitem[{Allen and Seidenberg(1999)}]{allen1999:emergence}
Joseph Allen and Mark~S Seidenberg. 1999.
\newblock The emergence of grammaticality in connectionist networks.
\newblock \emph{The emergence of language}, pages 115--151.

\bibitem[{Baevski et~al.(2019)Baevski, Auli, and
  Mohamed}]{baevski2019effectiveness}
Alexei Baevski, Michael Auli, and Abdelrahman Mohamed. 2019.
\newblock Effectiveness of self-supervised pre-training for speech recognition.
\newblock \emph{arXiv preprint arXiv:1911.03912}.

\bibitem[{Baevski et~al.(2020{\natexlab{a}})Baevski, Schneider, and
  Auli}]{Baevski2020vq-wav2vec}
Alexei Baevski, Steffen Schneider, and Michael Auli. 2020{\natexlab{a}}.
\newblock \href {https://openreview.net/forum?id=rylwJxrYDS} {vq-wav2vec:
  Self-supervised learning of discrete speech representations}.
\newblock In \emph{International Conference on Learning Representations}.

\bibitem[{Baevski et~al.(2020{\natexlab{b}})Baevski, Zhou, Mohamed, and
  Auli}]{baevski2020wav2vec}
Alexei Baevski, Henry Zhou, Abdelrahman Mohamed, and Michael Auli.
  2020{\natexlab{b}}.
\newblock wav2vec 2.0: A framework for self-supervised learning of speech
  representations.
\newblock \emph{arXiv preprint arXiv:2006.11477}.

\bibitem[{Baker et~al.(2014)Baker, Reichart, and
  Korhonen}]{baker2014unsupervised}
Simon Baker, Roi Reichart, and Anna Korhonen. 2014.
\newblock An unsupervised model for instance level subcategorization
  acquisition.
\newblock In \emph{Proceedings of the 2014 Conference on Empirical Methods in
  Natural Language Processing (EMNLP)}, pages 278--289.

\bibitem[{Bruni et~al.(2012)Bruni, Boleda, Baroni, and
  Tran}]{bruni2012distributional}
Elia Bruni, Gemma Boleda, Marco Baroni, and Nam-Khanh Tran. 2012.
\newblock Distributional semantics in technicolor.
\newblock In \emph{Proceedings of the 50th Annual Meeting of the Association
  for Computational Linguistics (Volume 1: Long Papers)}, pages 136--145.

\bibitem[{Chung and Glass(2018)}]{chung2018speech2vec}
Yu-An Chung and James Glass. 2018.
\newblock Speech2vec: A sequence-to-sequence framework for learning word
  embeddings from speech.
\newblock \emph{arXiv preprint arXiv:1803.08976}.

\bibitem[{Chung and Glass(2019)}]{chung2019generative}
Yu-An Chung and James Glass. 2019.
\newblock Generative pre-training for speech with autoregressive predictive
  coding.
\newblock \emph{arXiv preprint arXiv:1910.12607}.

\bibitem[{Devlin et~al.(2019)Devlin, Chang, Lee, and
  Toutanova}]{devlin2018bert}
Jacob Devlin, Ming-Wei Chang, Kenton Lee, and Kristina Toutanova. 2019.
\newblock {BERT}: Pre-training of deep bidirectional transformers for language
  understanding.
\newblock \emph{NAACL}.

\bibitem[{Dunbar et~al.(2019)Dunbar, Algayres, Karadayi, Bernard, Benjumea,
  Cao, Miskic, Dugrain, Ondel, Black, Besacier, Sakti, and
  Dupoux}]{dunbar2019zero}
Ewan Dunbar, Robin Algayres, Julien Karadayi, Mathieu Bernard, Juan Benjumea,
  Xuan-Nga Cao, Lucie Miskic, Charlotte Dugrain, Lucas Ondel, Alan~W. Black,
  Laurent Besacier, Sakriani Sakti, and Emmanuel Dupoux. 2019.
\newblock \href {http://arxiv.org/abs/1904.11469} {The zero resource speech
  challenge 2019: Tts without t}.

\bibitem[{Dunbar et~al.(2017)Dunbar, Cao, Benjumea, Karadayi, Bernard,
  Besacier, Anguera, and Dupoux}]{dunbar2017zero}
Ewan Dunbar, Xuan~Nga Cao, Juan Benjumea, Julien Karadayi, Mathieu Bernard,
  Laurent Besacier, Xavier Anguera, and Emmanuel Dupoux. 2017.
\newblock \href {http://arxiv.org/abs/1712.04313} {The zero resource speech
  challenge 2017}.

\bibitem[{Dunbar et~al.(2020)Dunbar, Karadayi, Bernard, Cao, Algayres, Ondel,
  Besacier, Sakriani, and Dupoux}]{dunbar2020}
Ewan Dunbar, Julien Karadayi, Mathieu Bernard, Xuan-Nga Cao, Robin Algayres,
  Lucas Ondel, Laurent Besacier, Sakti Sakriani, and Emmanuel Dupoux. 2020.
\newblock The zero resource speech challenge 2020: Discovering discrete subword
  and word units.
\newblock In \emph{{INTERSPEECH},
  perception;bootstrapping/modeling;clustering/bootphon}.

\bibitem[{Faruqui et~al.(2016)Faruqui, Tsvetkov, Rastogi, and
  Dyer}]{faruqui2016problems}
Manaal Faruqui, Yulia Tsvetkov, Pushpendre Rastogi, and Chris Dyer. 2016.
\newblock Problems with evaluation of word embeddings using word similarity
  tasks.
\newblock \emph{arXiv preprint arXiv:1605.02276}.

\bibitem[{Gerz et~al.(2016)Gerz, Vuli{\'c}, Hill, Reichart, and
  Korhonen}]{gerz2016simverb}
Daniela Gerz, Ivan Vuli{\'c}, Felix Hill, Roi Reichart, and Anna Korhonen.
  2016.
\newblock Simverb-3500: A large-scale evaluation set of verb similarity.
\newblock \emph{arXiv preprint arXiv:1608.00869}.

\bibitem[{Godais et~al.(2017)Godais, Linzen, and Dupoux}]{godais2017charnlm}
Gaël Godais, Tal Linzen, and Emmanuel Dupoux. 2017.
\newblock \href {https://doi.org/10.18653/v1/E17-2020} {Comparing
  character-level neural language models using a lexical decision task}.
\newblock pages 125--130.

\bibitem[{Goldberg(2019)}]{goldberg:2019}
Yoav Goldberg. 2019.
\newblock Assessing bert's syntactic abilities.
\newblock \emph{arXiv preprint 1901.05287}.

\bibitem[{Gulordava et~al.(2018)Gulordava, Bojanowski, Grave, Linzen, and
  Baroni}]{gulordava:2018}
Kristina Gulordava, Piotr Bojanowski, Edouard Grave, Tal Linzen, and Marco
  Baroni. 2018.
\newblock \href {https://www.aclweb.org/anthology/N18-1108} {Colorless green
  recurrent networks dream hierarchically}.

\bibitem[{Hahn and Baroni(2019)}]{hahn2019}
Michael Hahn and Marco Baroni. 2019.
\newblock \href {https://arxiv.org/abs/1906.07285} {Tabula nearly rasa: Probing
  the linguistic knowledge of character-level neural language models trained on
  unsegmented text}.
\newblock \emph{Transactions of the Association for Computational Linguistics
  (Accepted)}.

\bibitem[{Halawi et~al.(2012)Halawi, Dror, Gabrilovich, and
  Koren}]{halawi2012large}
Guy Halawi, Gideon Dror, Evgeniy Gabrilovich, and Yehuda Koren. 2012.
\newblock Large-scale learning of word relatedness with constraints.
\newblock In \emph{Proceedings of the 18th ACM SIGKDD international conference
  on Knowledge discovery and data mining}, pages 1406--1414.

\bibitem[{Hill et~al.(2015)Hill, Reichart, and Korhonen}]{hill2015simlex}
Felix Hill, Roi Reichart, and Anna Korhonen. 2015.
\newblock Simlex-999: Evaluating semantic models with (genuine) similarity
  estimation.
\newblock \emph{Computational Linguistics}, 41(4):665--695.

\bibitem[{Kahn et~al.(2020)Kahn, Riviere, Zheng, Kharitonov, Xu, Mazare,
  Karadayi, Liptchinsky, Collobert, Fuegen, and et~al.}]{Kahn_2020_librilight}
J.~Kahn, M.~Riviere, W.~Zheng, E.~Kharitonov, Q.~Xu, P.E. Mazare, J.~Karadayi,
  V.~Liptchinsky, R.~Collobert, C.~Fuegen, and et~al. 2020.
\newblock \href {https://doi.org/10.1109/icassp40776.2020.9052942}
  {Libri-light: A benchmark for asr with limited or no supervision}.
\newblock \emph{ICASSP 2020 - 2020 IEEE International Conference on Acoustics,
  Speech and Signal Processing (ICASSP)}.

\bibitem[{Kawakami et~al.(2020)Kawakami, Wang, Dyer, Blunsom, and van~den
  Oord}]{kawakami2020}
K.~Kawakami, L.~Wang, C.~Dyer, P.~Blunsom, and A.~van~den Oord. 2020.
\newblock \href {http://arxiv.org/abs/2001.11128} {Learning robust and
  multilingual speech representations}.

\bibitem[{Keuleers and Brysbaert(2010)}]{keuleers2010wuggy}
Emmanuel Keuleers and Marc Brysbaert. 2010.
\newblock Wuggy: A multilingual pseudoword generator.
\newblock \emph{Behavior research methods}, 42(3):627--633.

\bibitem[{Linzen et~al.(2016)Linzen, Dupoux, and Goldberg}]{linzen:2016}
Tal Linzen, Emmanuel Dupoux, and Yoav Goldberg. 2016.
\newblock Assessing the ability of {LSTMs} to learn syntax-sensitive
  dependencies.
\newblock \emph{TACL}.

\bibitem[{Liu et~al.(2019)Liu, Ott, Goyal, Du, Joshi, Chen, Levy, Lewis,
  Zettlemoyer, and Stoyanov}]{Liu2019RoBERTa}
Yinhan Liu, Myle Ott, Naman Goyal, Jingfei Du, Mandar Joshi, Danqi Chen, Omer
  Levy, Mike Lewis, Luke Zettlemoyer, and Veselin Stoyanov. 2019.
\newblock \href {http://arxiv.org/abs/1907.11692} {Roberta: {A} robustly
  optimized {BERT} pretraining approach}.
\newblock \emph{CoRR}, abs/1907.11692.

\bibitem[{Ludusan et~al.(2014)Ludusan, Versteegh, Jansen, Gravier, Cao,
  Johnson, and Dupoux}]{ludusan2014bridging}
Bogdan Ludusan, Maarten Versteegh, Aren Jansen, Guillaume Gravier, Xuan-Nga
  Cao, Mark Johnson, and Emmanuel Dupoux. 2014.
\newblock Bridging the gap between speech technology and natural language
  processing: an evaluation toolbox for term discovery systems.
\newblock In \emph{Proceedings of LREC}, pages 560--567.

\bibitem[{Luong et~al.(2013)Luong, Socher, and Manning}]{luong2013better}
Minh-Thang Luong, Richard Socher, and Christopher~D Manning. 2013.
\newblock Better word representations with recursive neural networks for
  morphology.
\newblock In \emph{Proceedings of the Seventeenth Conference on Computational
  Natural Language Learning}, pages 104--113.

\bibitem[{Miller and Charles(1991)}]{miller1991contextual}
George~A Miller and Walter~G Charles. 1991.
\newblock Contextual correlates of semantic similarity.
\newblock \emph{Language and cognitive processes}, 6(1):1--28.

\bibitem[{van~den Oord et~al.(2018)van~den Oord, Li, and Vinyals}]{oord2018cpc}
A{\"{a}}ron van~den Oord, Yazhe Li, and Oriol Vinyals. 2018.
\newblock \href {http://arxiv.org/abs/1807.03748} {Representation learning with
  contrastive predictive coding}.
\newblock \emph{CoRR}, abs/1807.03748.

\bibitem[{Ott et~al.(2019)Ott, Edunov, Baevski, Fan, Gross, Ng, Grangier, and
  Auli}]{ott-etal-2019-fairseq}
Myle Ott, Sergey Edunov, Alexei Baevski, Angela Fan, Sam Gross, Nathan Ng,
  David Grangier, and Michael Auli. 2019.
\newblock \href {https://doi.org/10.18653/v1/N19-4009} {fairseq: A fast,
  extensible toolkit for sequence modeling}.
\newblock In \emph{Proceedings of the 2019 Conference of the North {A}merican
  Chapter of the Association for Computational Linguistics (Demonstrations)},
  pages 48--53, Minneapolis, Minnesota. Association for Computational
  Linguistics.

\bibitem[{{Panayotov} et~al.(2015){Panayotov}, {Chen}, {Povey}, and
  {Khudanpur}}]{Panayotov2015librispeech}
V.~{Panayotov}, G.~{Chen}, D.~{Povey}, and S.~{Khudanpur}. 2015.
\newblock Librispeech: An asr corpus based on public domain audio books.
\newblock In \emph{2015 IEEE International Conference on Acoustics, Speech and
  Signal Processing (ICASSP)}, pages 5206--5210.

\bibitem[{Radinsky et~al.(2011)Radinsky, Agichtein, Gabrilovich, and
  Markovitch}]{radinsky2011word}
Kira Radinsky, Eugene Agichtein, Evgeniy Gabrilovich, and Shaul Markovitch.
  2011.
\newblock A word at a time: computing word relatedness using temporal semantic
  analysis.
\newblock In \emph{Proceedings of the 20th international conference on World
  wide web}, pages 337--346.

\bibitem[{Ravfogel et~al.(2018)Ravfogel, Tyers, and Goldberg}]{ravfogel2018can}
Shauli Ravfogel, Francis~M Tyers, and Yoav Goldberg. 2018.
\newblock Can {LSTM} learn to capture agreement? the case of basque.
\newblock \emph{arXiv preprint 1809.04022}.

\bibitem[{Rivière et~al.(2020)Rivière, Joulin, Mazaré, and
  Dupoux}]{riviere2020unsupervised}
Morgane Rivière, Armand Joulin, Pierre-Emmanuel Mazaré, and Emmanuel Dupoux.
  2020.
\newblock \href {http://arxiv.org/abs/2002.02848} {Unsupervised pretraining
  transfers well across languages}.

\bibitem[{Rubenstein and Goodenough(1965)}]{rubenstein1965contextual}
Herbert Rubenstein and John~B Goodenough. 1965.
\newblock Contextual correlates of synonymy.
\newblock \emph{Communications of the ACM}, 8(10):627--633.

\bibitem[{Salazar et~al.(2020)Salazar, Liang, Nguyen, and
  Kirchhoff}]{salazar-etal-2020-masked}
Julian Salazar, Davis Liang, Toan~Q. Nguyen, and Katrin Kirchhoff. 2020.
\newblock \href {https://doi.org/10.18653/v1/2020.acl-main.240} {Masked
  language model scoring}.
\newblock In \emph{Proceedings of the 58th Annual Meeting of the Association
  for Computational Linguistics}, pages 2699--2712, Online. Association for
  Computational Linguistics.

\bibitem[{Schatz et~al.(2013)Schatz, Peddinti, Bach, Jansen, Hermansky, and
  Dupoux}]{Schatz2013ABX}
T.~Schatz, V.~Peddinti, F.~Bach, A.~Jansen, H.~Hermansky, and E.~Dupoux. 2013.
\newblock Evaluating speech features with the minimal-pair abx task: Analysis
  of the classical mfc/plp pipeline.
\newblock \emph{INTERSPEECH}.

\bibitem[{Schatz(2016)}]{schatz2016abx}
Thomas Schatz. 2016.
\newblock \emph{ABX-discriminability measures and applications}.
\newblock Ph.D. thesis, Paris 6.

\bibitem[{Schnabel et~al.(2015)Schnabel, Labutov, Mimno, and
  Joachims}]{schnabel2015evaluation}
Tobias Schnabel, Igor Labutov, David Mimno, and Thorsten Joachims. 2015.
\newblock Evaluation methods for unsupervised word embeddings.
\newblock In \emph{Proceedings of the 2015 conference on empirical methods in
  natural language processing}, pages 298--307.

\bibitem[{Schneider et~al.(2019)Schneider, Baevski, Collobert, and
  Auli}]{schneider2019wav2vec}
S.~Schneider, A.~Baevski, R.~Collobert, and M.~Auli. 2019.
\newblock wav2vec: Unsupervised pre-training for speech recognition.
\newblock \emph{arXiv:1904.05862}.

\bibitem[{Vaswani et~al.(2017)Vaswani, Shazeer, Parmar, Uszkoreit, Jones,
  Gomez, Kaiser, and Polosukhin}]{vaswani2017attention}
Ashish Vaswani, Noam Shazeer, Niki Parmar, Jakob Uszkoreit, Llion Jones,
  Aidan~N. Gomez, Lukasz Kaiser, and Illia Polosukhin. 2017.
\newblock \href {http://arxiv.org/abs/1706.03762} {Attention is all you need}.
\newblock \emph{CoRR}, abs/1706.03762.

\bibitem[{Versteegh et~al.(2016)Versteegh, Anguera, Jansen, and
  Dupoux}]{versteegh2015zero}
Maarten Versteegh, Xavier Anguera, Aren Jansen, and Emmanuel Dupoux. 2016.
\newblock \href {https://doi.org/10.1016/j.procs.2016.04.031} {The zero
  resource speech challenge 2015: Proposed approaches and results}.
\newblock \emph{Procedia Computer Science}, 81:67--72.

\bibitem[{Wang et~al.(2020)Wang, Tang, and Livescu}]{wang2020}
Weiran Wang, Qingming Tang, and Karen Livescu. 2020.
\newblock Unsupervised pre-training of bidirectional speech encoders via masked
  reconstruction.
\newblock In \emph{ICASSP 2020-2020 IEEE International Conference on Acoustics,
  Speech and Signal Processing (ICASSP)}, pages 6889--6893. IEEE.

\bibitem[{Warstadt et~al.(2019)Warstadt, Parrish, Liu, Mohananey, Peng, Wang,
  and Bowman}]{warstadt2019blimp}
Alex Warstadt, Alicia Parrish, Haokun Liu, Anhad Mohananey, Wei Peng, Sheng-Fu
  Wang, and Samuel~R Bowman. 2019.
\newblock Blimp: A benchmark of linguistic minimal pairs for english.
\newblock \emph{arXiv preprint arXiv:1912.00582}.

\bibitem[{Yang and Powers(2006)}]{yang2006verb}
Dongqiang Yang and David~Martin Powers. 2006.
\newblock \emph{Verb similarity on the taxonomy of WordNet}.
\newblock Masaryk University.

\end{thebibliography}
\bibliographystyle{acl_natbib}

\appendix

\clearpage
\begin{center}
\textbf{\large Supplementary Materials}
\end{center}
\setcounter{equation}{0}
\setcounter{figure}{0}
\setcounter{table}{0}
\setcounter{page}{1}
\makeatletter
\renewcommand{\theequation}{S\arabic{equation}}
\renewcommand{\thefigure}{S\arabic{figure}}
\renewcommand{\thetable}{S\arabic{table}}
\renewcommand{\bibnumfmt}[1]{[S#1]}
\renewcommand{\citenumfont}[1]{S#1}

\section{Sampling method to balance ngram scores}\label{SM:sampling}
We describe here our sampling method to balance ngram scores for sWUGGY and sBLIMP datasets.
We first show the algorithm that we applied to sWUGGY, then we just modify slightly the algorithm for the sBLIMP dataset.

For sWUGGY, let's assume that we have $N$ words $w_1,\dots, w_N$; and for each word $w_i$, we have a list of $K$ matching nonword candidates $nw_i^1,\dots, nw_i^K$. We also assume that each word or nonword $w$ has $M$ scores $s_1(w),\dots,s_M(w)$ (this might be unigram/bigram char/phone scores). We aim to choose, for each word $w_i$, a matching nonword $nw_i^*$ such that the proportion of the pairs where the score of the word is higher than the score of nonword is close to 50\% as possible, for each of $M$ scores.

In other words, we want to build a list of word-nonword pairs $L=\{(w_1, nw_1^*),\dots,(w_N, nw_N^*)\}$ such that the objective function
\begin{equation}\label{eq:scoreObj}
    \text{obj}(L) = \sum_{m=1}^M\lvert\text{accuracy\_of\_score\_m(L)-0.5}\rvert
\end{equation}
is as close to zero as possible.

We thus deduce a simple sampling method as follows: We first initialize a list $L$ of chosen pairs of word and nonword. At each iteration, we randomly choose an unchosen word. Then we sample a nonword candidate in the list of matching nonword candidates, update the list with the new pair, and compute the objective function of the new list as given in \ref{eq:scoreObj}. If the objective increases, we remove this newly added element, and resample a new nonword from the list of candidates. If we encounter all the nonword candidates but cannot find a new pair, we randomly choose a nonword from the list of candidates. We then continue to the next word until all the words are chosen.

We found afterwards that if we sample all the words at the same time, we can obtain an overall score very close to 50\%, but then words with high frequency or with short length tended to have higher accuracy than others. We then decided to divide the words into sub-categories by frequency and word length, and then do the sampling on each of the sub-categories, which gives a more balanced score on all the length and frequency levels.

For sBLIMP, the candidates are slightly different. We now have a list of $N$ pairs of grammatical and non-grammatical sentences and we want to choose $K$ pairs among them such that the accuracy of the chosen pairs is as close to 50\% as possible as for sWUGGY.
We can then use the same sampling method as described above, with the exception that instead of choosing a word and sampling the nonword candidates at each iteration, we sample an unchosen pair in the list of candidates, and add that pair to the chosen list if we succeed to decrease the objective function.

As we also found that there is a huge difference in the accuracy scores of linguistic paradigms, we tried to do the sampling by each sub-paradigm. However, there were still some paradigms for which we were not able to perfectly balance the score.




\section{Supplementary ABX methods and results}

Given two sounds $x$ and $y$ with two sequences of representations $\textbf{r}^x = r^x_1,\dots,r^x_T$ and
$\textbf{r}^y = r^y_1,\dots,r^y_S$ respectively, the ABX distance between $x$ and $y$ is computed as follows: 
\begin{equation}\label{eq:ABX}
    d_{ABX}(x,y) = \frac{1}{\lvert\text{path}_{\text{DTW}}(\textbf{r}^x,\textbf{r}^y)\rvert}\sum_{(i,j)\in\text{path}_{\text{DTW}}(\textbf{r}^x,\textbf{r}^y)} sim(r^x_i,r^y_j).
\end{equation}

where $sim(x,y)$ is the arc cosine of the normalized dot product between the embeddings $x$ and $y$.

Table \ref{tab:gruABX} shows the ABX error on Libri-light dev-clean as a function of different hidden layer of the autoregressive network. We found that as long as we have a big autoregressive network, it is generally not the last layer that brings the best phonetic information of the audio file.
\begin{table}[h]
\caption{Within and Across Speaker ABX error (lower is better) on Libri-light dev-clean at different level of the autoregressive network of CPC-small and CPC-big models. Best layer for each model in bold.}
\label{tab:gruABX}
\begin{center}
\begin{tabular}{lcclcccc}
\thickhline
\rowcolor{white}
 & \multicolumn{2}{c}{CPC-small} & & \multicolumn{4}{c}{CPC-big}  \\
 \cline{2-3} \cline{5-8}
\rule{0pt}{2ex} LSTM layer & 1  & 2   & & 1 & 2 & 3 & 4 \\
\thickhline
\rule{0pt}{2ex}within & 10.26 &\bf	6.24 &&	9.62&\bf	3.41&	4.65&	9.50  \\
    across & 14.17&\bf	8.17&&	14.73&\bf	4.18&	5.40&	9.95 \\  
\thickhline
\end{tabular}
\end{center}

\end{table}

Table \ref{tab:kmeansABX} reports the ABX scores for different number of clusters, we also included multiple-group clustering in our experiences as similar to \cite{Baevski2020vq-wav2vec}. We found that the best score is obtained with 50 clusters. Using multiple groups do not further improve the quality of the discretized units, this may be due to the fact that we only used one-hot information of the multiple groups (for example, the two codes 26-20 and 26-10 represent two different one-hot units without any correlation).
\begin{table}[h]
\caption{Within and Across-Speaker ABX error rate (lower is better) on the LibriSpeech dev-clean dataset for CPC-small+kmeans (one-hot vectors embeddings) with different number of units (clusterings). Optimal number of clusters in bold.}
\label{tab:kmeansABX}
\begin{center}
    \begin{tabular}{rccccccc}
    \thickhline
     \rule{0pt}{2ex} nunits & 20     & 50     & 200    & 500    & 2000   & \multicolumn{1}{l}{50 x 2gr} & \multicolumn{1}{l}{320 x 2gr} \\
    \thickhline
    \rule{0pt}{2ex}within & 11.3 &\bf 10.3 & 12.5 & 13.4 & 17.0 & 12.6 & 18.3  \\
    across & 14.5 &\bf 14.2 & 16.8 & 19.9 & 27.2 & 17.7 & 27.7 \\  
    \thickhline
    \end{tabular}
\end{center}
\end{table}

\section{Supplementary spot-the-word results}

\begin{table}[h]
\caption{\textbf{Spot-the-word accuracy} (higher is better) on sWUGGY dev as a function of the masking parameters to compute the pseudo-probabilities. The runtime is estimated based on the evaluation time with the base parameters $M_d=\Delta t = 10$. In bold the compromise we selected between accuracy and speed.}
\label{tab:masking}
\vspace{-0.5em}
\resizebox{\textwidth}{!}{
\begin{tabular}{r|cc|ccc|ccc|ccc}
\thickhline
\rule{0pt}{2ex} $M_d$ & \multicolumn{2}{c|}{5}  & \multicolumn{3}{c|}{10} & \multicolumn{3}{c|}{15} & \multicolumn{3}{c}{20}     \\
\hline
\rule{0pt}{2ex} $\Delta t$  & 5  & 1        & 10     & 5      & 1           & 15       & 5    & 1       & 20   & 5    & 1       \\
\thickhline
\rule{0pt}{2ex} scores  & 59.14 & 62.59  & 64.59 & 68.23 & 70.85 & 66.45 & \textbf{70.69}  & 72.52  & 64.38    & 69.04 & 71.33  \\
runtime (est.) & $\times$ 2 & $\times$ 10 & $\times$ 1 & $\times$ 2 & $\times$ 10 & $\times$ 0.66 & \textbf{$\times$ 2} & $\times$ 10 & $\times$ 0.5 & $\times$ 2 & $\times$ 10 \\
\thickhline
\end{tabular}}
\end{table}

Table \ref{tab:masking} investigates the effect of the masking parameters $M_d$ and $\Delta t$ to the spot-the-word metrics. We found that the way of computing log-probability can greatly influence the evaluation scores. We see that as long as we overlap the masking spans more, the performance is better. In addition, given that we masked spans of $M\sim \mathcal{N}(10,10)$ tokens during training, the best decoding masking size was found to be $15$. Considering the evaluation time, it is theoretically inversely proportional to $\Delta t$, and we thus decided to choose $M_d=15$ and $\Delta t=5$ for an accuracy and speed trade-off.

\begin{figure*}[]
\centering
{\includegraphics[width=0.98\textwidth]{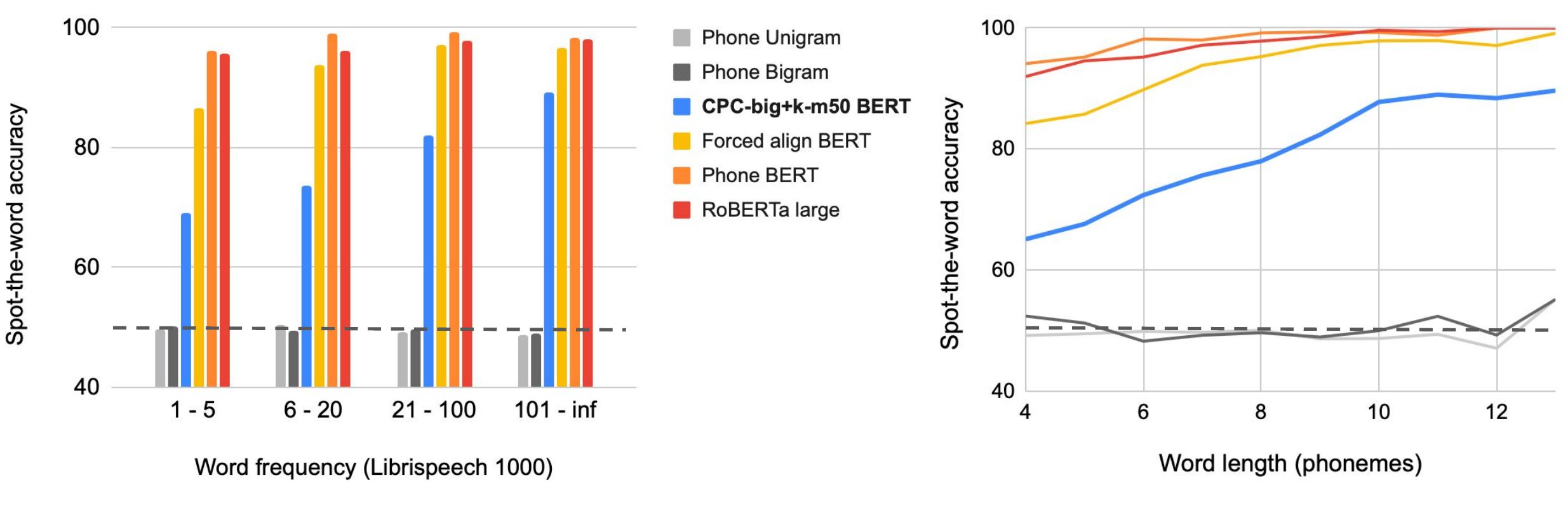}}
\vspace{-.7em}
\caption{{\bf Spot-the-word accuracy} (sWUGGY dev set, higher is better, chance level at 50\%) for our best CPC+clustering+BERT model (blue), compared to phone ngram baselines (gray) and text-based transformer toplines (orange). Left, word frequency effect. Right, word length effect.}
\label{fig:WUGGY}
\end{figure*}

Figure \ref{fig:WUGGY} shows the performance of the CPC-big system on the BERT-large architecture: they are worse than the toplines but well above chance. We reproduce the frequency effects (more frequent words giving rise to better accuracies) and the length effect (longer words giving rise to better accuracies).
This may be due to the fact that the phonetic space is sparser for long than for short words. As a consequence, a short nonword like "tup" could be continued as a real word in multiple ways ("tuple", "tupperware", etc.)
. In contrast, a long nonword can rarely be salvaged into a word (eg, 'rhanoceros' is a nonword very early on). 

\section{Supplementary grammaticality results}
Table \ref{tab:BLIMP} shows the detailed results on the various subsets of sBLIMP of our best model. Almost all of the subsets show better than chance scores (11/12), and of the phoneme ngrams controls (11/12), and most are better than the word ngrams controls (9/12 for unigram models, and 10/12 for bigram models). 

\begin{table*}[h!]
    \caption{{\bf Sentence acceptability accuracy} (sBLIMP dev set, higher is better, chance level at 50\%) for our best CPC+kmeans 50+BERT model, compared to phone ngram baselines, text-based transformer toplines, and human scores \citep[from][]{warstadt2019blimp}.}
    \vspace{-1em}
    \label{tab:BLIMP}
    \begin{center}
    \resizebox{\textwidth}{!}{
        \addtolength\tabcolsep{2pt}
        \begin{tabular}{r*{13}{c}}
        \hline
          & \mcrot{1}{c}{45}{\textbf{Overall}}    & \mcrot{1}{c}{45}{Ana. Agr.}    & \mcrot{1}{c}{45}{Agr. Str.}    & \mcrot{1}{c}{45}{Binding}    & \mcrot{1}{c}{45}{Ctrl. Rais.}    & \mcrot{1}{c}{45}{D-N Agr.}    & \mcrot{1}{c}{45}{Ellipsis}    & \mcrot{1}{c}{45}{Fill. Gap.}    & \mcrot{1}{c}{45}{Irregular}    & \mcrot{1}{c}{45}{Island} & \mcrot{1}{c}{45}{NPI Li.} & \mcrot{1}{c}{45}{Quantifiers} & \mcrot{1}{c}{45}{S-V Arg.} \\ \hline
          \T Phone Unigram \ \ \B &  \textbf{\gradientBLIMP{48.29}} & \gradientBLIMP{50.00} & \gradientBLIMP{50.00} & \gradientBLIMP{52.90} & \gradientBLIMP{50.00} & \gradientBLIMP{50.00} & \gradientBLIMP{50.00} & \gradientBLIMP{50.00} & \gradientBLIMP{45.50} & \gradientBLIMP{50.00} & \gradientBLIMP{38.36} & \gradientBLIMP{39.33} & \gradientBLIMP{50.00} \\
        \T Phone Bigram \ \ \B &  \textbf{\gradientBLIMP{50.20}} & \gradientBLIMP{50.50} & \gradientBLIMP{50.11} & \gradientBLIMP{52.40} & \gradientBLIMP{49.80} & \gradientBLIMP{50.12} & \gradientBLIMP{50.00} & \gradientBLIMP{49.88} & \gradientBLIMP{50.00} & \gradientBLIMP{49.93} & \gradientBLIMP{50.00} & \gradientBLIMP{50.00} & \gradientBLIMP{50.00} \\
        \T Word Unigram \ \ \B &  \textbf{\gradientBLIMP{54.40}} & \gradientBLIMP{50.50} & \gradientBLIMP{50.06} & \gradientBLIMP{65.20} & \gradientBLIMP{49.90} & \gradientBLIMP{50.06} & \gradientBLIMP{49.50} & \gradientBLIMP{75.00} & \gradientBLIMP{51.00} & \gradientBLIMP{50.00} & \gradientBLIMP{49.79} & \gradientBLIMP{50.00} & \gradientBLIMP{49.92} \\
        \T Word Bigram \ \ \B &  \textbf{\gradientBLIMP{51.64}} & \gradientBLIMP{50.00} & \gradientBLIMP{50.06} & \gradientBLIMP{66.50} & \gradientBLIMP{50.00} & \gradientBLIMP{50.06} & \gradientBLIMP{49.00} & \gradientBLIMP{50.00} & \gradientBLIMP{50.00} & \gradientBLIMP{50.07} & \gradientBLIMP{50.00} & \gradientBLIMP{57.00} & \gradientBLIMP{49.92} \\ 
        \hline
        \T CPC-big+km50 BERT \ \ \B &  \textbf{\gradientBLIMP{56.14}} & \gradientBLIMP{61.50} & \gradientBLIMP{51.10} & \gradientBLIMP{62.30} & \gradientBLIMP{51.62} & \gradientBLIMP{60.66} & \gradientBLIMP{74.75} & \gradientBLIMP{59.91} & \gradientBLIMP{55.44} & \gradientBLIMP{56.64} & \gradientBLIMP{48.29} & \gradientBLIMP{63.25} & \gradientBLIMP{51.62} \\
        \T Forced phone BERT \ \ \B &  \textbf{\gradientBLIMP{63.72}} & \gradientBLIMP{72.62} & \gradientBLIMP{56.40} & \gradientBLIMP{63.80} & \gradientBLIMP{54.90} & \gradientBLIMP{80.47} & \gradientBLIMP{69.00} & \gradientBLIMP{66.34} & \gradientBLIMP{79.94} & \gradientBLIMP{58.71} & \gradientBLIMP{54.29} & \gradientBLIMP{61.00} & \gradientBLIMP{65.12} \\
        \T Phone BERT \ \ \B &  \textbf{\gradientBLIMP{66.78}} & \gradientBLIMP{72.50} & \gradientBLIMP{59.89} & \gradientBLIMP{54.40} & \gradientBLIMP{62.20} & \gradientBLIMP{92.25} & \gradientBLIMP{75.00} & \gradientBLIMP{63.75} & \gradientBLIMP{82.50} & \gradientBLIMP{57.71} & \gradientBLIMP{54.57} & \gradientBLIMP{81.67} & \gradientBLIMP{70.17} \\
        \hline
        \T RoBERTa large \ \ \B &  \textbf{\gradientBLIMP{81.56}} & \gradientBLIMP{98.50} & \gradientBLIMP{74.33} & \gradientBLIMP{80.40} & \gradientBLIMP{78.20} & \gradientBLIMP{95.88} & \gradientBLIMP{99.00} & \gradientBLIMP{73.62} & \gradientBLIMP{89.50} & \gradientBLIMP{68.71} & \gradientBLIMP{80.71} & \gradientBLIMP{90.67} & \gradientBLIMP{87.83} \\
        \T Human (on BLIMP original) \ \ \B &  \textbf{\gradientBLIMP{88.60}} & \gradientBLIMP{97.50} & \gradientBLIMP{90.00} & \gradientBLIMP{87.30} & \gradientBLIMP{83.90} & \gradientBLIMP{92.20} & \gradientBLIMP{85.00} & \gradientBLIMP{86.90} & \gradientBLIMP{97.00} & \gradientBLIMP{84.90} & \gradientBLIMP{88.10} & \gradientBLIMP{86.60} & \gradientBLIMP{90.90} \\ \hline
           
        \end{tabular}}
    \end{center}

\vspace{-1.5em}
\end{table*}
\section{Supplementary semantic similarity results}
Table \ref{tab:ABX-and-SIMI} shows the detailed sSIMI results, layer by layer of the best BERT model together with the detailed ABX results on the same layers. This shows a complementarity of these two metrics (the best layers for acoustics/phonetics are the worst for semantics and vice versa).
\vspace{-1em}
\begin{table*}[h]
    \caption{Comparison of \textbf{Semantic similarity scores} (Spearman's correlation with human judgement, higher is better) on the sSIMI synthetic dev set and \textbf{ABX scores} on Libri-light dev-clean on different embedding levels of our CPC-big+kmeans50+BERT model. \textit{CPC} refers to the outputs of the second LSTM hidden layer of the CPC-big model, \textit{kmeans} and \textit{outs} refers to 1-hot representations before and after the BERT model repsectively. The semantic similarity scores are also evaluated with different pooling function (mean, max, min). Higher error rates than MFCC baseline in ABX and negative SIMI scores are in red.
    Note that all the semantic similarity scores are multiplied by 100.}
    \label{tab:ABX-and-SIMI}
    \vspace{-1em}
    \begin{center}
    
    \resizebox{\textwidth}{!}{
        \setlength\tabcolsep{0pt}
        \begin{tabular}{rr*{17}{c}}
        \hline
           & \multirow{2}{*}{\ Score \ } & \multirow{2}{*}{\ CPC \ } & \multirow{2}{*}{\ kmeans \ } & \multicolumn{15}{c}{BERT Layer}\\
           & &   &  & 0 & 1 & 2 & 3 & 4 & 5 & 6 & 7 & 8 & 9 & 10 & 11 & 12 & \ logits \ & \ outs \ \\
          \hline
          \multirow{2}{*}[-1.5ex]{ABX \ } & \T within \ \B & \gradientBoxABXw{3.41} & \gradientBoxABXw{6.38} & \gradientBoxABXw{11.82} & \gradientBoxABXw{21.97} & \gradientBoxABXw{35.02} & \gradientBoxABXw{42.54} & \gradientBoxABXw{47.40} & \gradientBoxABXw{44.46} & \gradientBoxABXw{43.71} & \gradientBoxABXw{41.73} & \gradientBoxABXw{33.76} & \gradientBoxABXw{19.67} & \gradientBoxABXw{15.91} & \gradientBoxABXw{15.93} & \gradientBoxABXw{3.30} & \gradientBoxABXw{3.65} & \gradientBoxABXw{5.65} \\
        & \T across \ \B & \gradientBoxABXa{4.18} & \gradientBoxABXa{8.26} & \gradientBoxABXa{13.77} & \gradientBoxABXa{24.59} & \gradientBoxABXa{36.95} & \gradientBoxABXa{43.90} & \gradientBoxABXa{47.94} & \gradientBoxABXa{45.52} & \gradientBoxABXa{44.76} & \gradientBoxABXa{43.12} & \gradientBoxABXa{36.29} & \gradientBoxABXa{23.13} & \gradientBoxABXa{18.92} & \gradientBoxABXa{18.84} & \gradientBoxABXa{4.11} & \gradientBoxABXa{4.59} & \gradientBoxABXa{7.32} \\ 
        \hline
         \multirow{3}{*}[-2ex]{\ sSIMI \ } &  \T mean \ \B & - & - & \gradientBox{-0.58} & \gradientBox{-1.97} & \gradientBox{-1.54} & \gradientBox{0} & \gradientBox{1.47} & \gradientBox{-0.38} & \gradientBox{1.04} & \gradientBox{2.26} & \gradientBox{1.71} & \gradientBox{2.26} & \gradientBox{1.47} & \gradientBox{2.96} & \gradientBox{-0.57} & - & - \\
        & \T max \ \B & - & - & \gradientBox{-1.79} & \gradientBox{0.25} & \gradientBox{0.51} & \gradientBox{5.02} & \gradientBox{6.25} & \gradientBox{4.03} & \gradientBox{2.61} & \gradientBox{1.86} & \gradientBox{1.69} & \gradientBox{0.83} & \gradientBox{1.78} & \gradientBox{1.78} & \gradientBox{0.09} & - & - \\
        & \T min \ \B & - & - & \gradientBox{-3.3} & \gradientBox{-1.12} & \gradientBox{-0.93} & \gradientBox{0.86} & \gradientBox{6.21} & \gradientBox{1.9} & \gradientBox{0.96} & \gradientBox{0.12} & \gradientBox{3.53} & \gradientBox{5.03} & \gradientBox{0.71} & \gradientBox{3.41} & \gradientBox{-0.9} & - & - \\\hline
        \end{tabular}}
    \end{center}

\vspace{-1.3em}
\end{table*}

\end{document}